\documentclass[runningheads]{llncs}
\usepackage{graphicx}
\usepackage{comment}

\usepackage[toc,page]{appendix}
\usepackage{subfigure}
\usepackage{diagbox}
\usepackage{amssymb}
\usepackage{pifont}
\newcommand{\cmark}{\ding{51}}%
\newcommand{\xmark}{\ding{55}}%
\usepackage{wrapfig}
\usepackage{graphicx}
\usepackage[utf8]{inputenc}
\usepackage{multirow}
\usepackage{xcolor}

\begin{document}
\title{Looking for COVID-19 misinformation in multilingual social media texts}
%
%
\author{ Raj Ratn Pranesh\inst{1} \and
Mehrdad Farokhnejad\inst{2} \\ \and
Ambesh Shekhar\inst{1} \and Genoveva Vargas-Solar\inst{4}}
\authorrunning{Raj Ratn Pranesh et al.}
%
\institute{Birla Institute of Technology, Mesra, India \\ \email{raj.ratn18@gmail.com,  ambesh.sinha@gmail.com}\and
Univ. Grenoble Alpes, CNRS, LIG, Grenoble, France \\
\email{Mehrdad.Farokhnejad@univ-grenoble-alpes.fr}\\  \and
CNRS, LIRIS-LAFMIA Lyon, France \\
\email{genoveva.vargas-solar@liris.cnrs.fr}}
\maketitle              
\begin{abstract}
This paper presents the Multilingual COVID-19  Analysis Method (CMTA)  for detecting and observing the spread of misinformation about this disease within texts.
CMTA proposes a data science (DS) pipeline that applies machine learning models for processing, classifying (Dense-CNN) and analyzing (MBERT) multilingual (micro)-texts. 
DS pipeline data preparation tasks extract features from multilingual textual data and categorize it into specific information classes (i.e., 'false', 'partly false', 'misleading').  
 The CMTA pipeline has been experimented with multilingual micro-texts (tweets), showing misinformation spread across different languages. To assess the performance of CMTA and put it in perspective, we performed a comparative analysis of CMTA with eight monolingual models used for detecting misinformation. The comparison shows that CMTA  has surpassed various monolingual models and suggests that it can be used as a general method for detecting misinformation in multilingual micro-texts. CMTA experimental results show misinformation trends about COVID-19 in different languages during the first pandemic months.

\end{abstract}
%
%
%

\section{Introduction}\label{sec:intro}

Since late 2019, the coronavirus disease COVID-19 has spread worldwide to more than 216 countries \cite{world2020coronavirus}. 
The COVID-19 pandemic has highlighted the extent to which the world's population is interconnected through the Internet and social media.
Indeed, social media is a significant conduit where people share their response, thoughts, news, information related to COVID-19, with one in three individuals worldwide participating in social media, with two-thirds of people utilizing it on the Internet \cite{Esteban2019,wilford2018social}. 
 Social media provides particularly fertile ground for the spread of information and misinformation \cite{frenkel2020surge}.  It can even give direct access to  content, which may intensify rumours and dubious information \cite{cinelli2020covid}. For people with non-medical experience it is difficult to assess health information's authenticity.
Misinformation may have intense implications for public opinion and behaviour, positively or negatively influencing the viewpoint of those who access it\cite{brindha2020social,kouzy2020coronavirus}. 
Seeking accurate and valid information is the biggest challenge with Internet health information during the pandemic \cite{eysenbach2002empirical}.

This paper proposes  CMTA, a multilingual tweet 
analysis and information (misinformation) detection method for observing  social media misinformation spread about the COVID-19 pandemic within communities with different languages. CMTA proposes a data science pipeline with tasks that rely on popular artificial intelligence models (e.g., multi-lingual BERT and CNN) to process texts and classifying them according to different misinformation classes ('false', 'partly false' and 'misleading').  
The paper describes the experimental setting implemented for validating CMTA that uses datasets of COVID-19 related tweets. 
The paper presents an illustrative statistical representation of the findings insisting on the insights discovered in our study to discuss results. To assess the performance of CMTA, 
we compared  CMTA with eight monolingual BERT models. The comparison shows that CMTA  has surpassed various monolingual models and suggests that it can be used as a general method for detecting misinformation in multilingual micro-texts.

The remainder of the paper is organised as follows. Section \ref{sec:related} introduces works that have addressed misinformation detection about COVID-19 on social media datasets. 
Section \ref{sec:model} describes the general approach behind the method  CMTA that we propose. 
It describes the experiment setting that we used for validating CMTA.    Section \ref{sec:assessment} compares the performance of CMTA against mono-lingual analytics performed on the same dataset.  It also discusses the study results about misinformation spread through micro-texts (i.e., tweets) across different languages. 
Finally, Section \ref{sec:conclusion} concludes the paper and discusses future work.

\section{Related work}\label{sec:related}
The COVID-19 pandemic has resulted in studies investigating the various types of misinformation arising during the COVID-19 crisis \cite{brennen2020types,dharawat2020drink,singh2020first,kouzy2020coronavirus}. Studies investigate a  small subset of claims \cite{singh2020first} or manually annotate Twitter data \cite{kouzy2020coronavirus}. In
\cite{brennen2020types} authors analyse  different types of
sources for looking for COVID-19 misinformation. Pennycook et al. \cite{pennycook2020fighting} introduced an attention-based account of misinformation and observed that people tend to believe false claims about COVID-19 and share false claims when they do not think critically about the accuracy and veracity of the information. Kouzy et al. \cite{kouzy2020coronavirus} annotated about 600 messages containing hashtags about COVID-19, they observed that about one-fourth of messages contain some form of misinformation, and about 17\% contain some unverifiable information. With such misinformation overload, any decision making procedure based on misinformation has a high likelihood of severely impacting people's health \cite{ingraham2020fact}.
The work in \cite{huang2020disinformation} examined the global spread of information related to crucial  disinformation stories and "fake news" URLs  during the early stages of the global pandemic on
Twitter.  Their study shows that news agencies, government officials, and individual news reporters send messages that spread widely and play critical roles. Tweets citing URLs for "fake news" and reports of propaganda are more likely than news or government pages shared by regular users and bots.

The work in \cite{sharma2020covid} focused on topic modelling and designed a dashboard to track Twitter's misinformation regarding the COVID-19 pandemic. The dashboard presents a summary of information derived from Twitter posts, including topics, sentiment, false and misleading information shared on social media related to COVID-19.
Cinelli et al. \cite{singh2020first} track (mis)-information flow across 2.7M tweets and compare it with infection rates.They noticed a major Spatio-temporal connection between information flow and new COVID-19 instances, and while there are discussions about myths and connections to low-quality information, their influence is less prominent than other themes specific to the crisis.
To find and measure causal relationships between pandemic features (e.g. the number of infections and deaths) and Twitter behaviour and public sentiment, the work in \cite{gencoglu2020causal} introduced the first example of a causal inference method. Their proposed approach has shown that they can efficiently collect epidemiological domain knowledge and identify factors that influence public interest and attention.

The discussion around the COVID-19 pandemic and the government policies was investigated in\cite{lopez2020understanding}. They used Twitter data in multiple languages from various countries and found common responses to the pandemic and how they differ across time using text mining. Moreover, they presented insights as to how information and misinformation were transmitted via Twitter. Similarly, to demonstrate the epidemiological effect of COVID-19 on press publications in Bogota, Colombia, \cite{saire2020people} used text mining on Twitter data. They intuitively note a strong correlation between the number of tweets and the number of infected people in the area.
 
Most of the works described above focus on analyzing tweets related to a single language such as English. Our work has designed a single model leveraging multilingual BERT for the analysis of tweets in multiple languages. Furthermore, we used a large data set to train and analyze the tweets.  We aim to provide a system that will not be restricted to any language for analyzing social media data.

\section{The CMTA Method}\label{sec:model}
Both misinformation\footnote { \url{https://www.oed.com/view/Entry/119699?redirectedFrom=misinformation}} and disinformation\footnote{ \url{https://www.oed.com/view/Entry/54579?redirectedFrom=disinformation}}, according to the Oxford English Dictionary, are false or misleading information. Misinformation refers to information that is accidentally false and spread without the intent to hurt, whereas disinformation refers to false information that is intentionally produced and shared to cause hurt \cite{hernon1995disinformation}.
Claims do not have to be entirely truthful or incorrect; they can contain a small amount of false or inaccurate  information\cite{shahi2020fakecovid}. 
This work uses the general notion of misinformation and makes no distinction between misinformation and disinformation as it is practically difficult to determine one's intention computationally.


The data science pipeline phases proposed by CMTA are: tokenizing, text features extraction, linear transformation, and classification. The first phases   (tokenizing, text feature extraction, linear transformation) correspond to a substantial data-preparation process intended to build a multi-lingual vectorized representation of texts. The objective is to achieve a numerical pivot representation of texts agnostic of the language.
CMTA classification task uses a dense layer and leads to a trained network model that can be used to classify micro-texts (e.g. tweets) into three misinformation classes: 'false', 'partly false' and 'misleading'. 

\vspace{-0,3cm}
\paragraph{Text tokenization}

Given a multilingual textual dataset consisting of sentences, CMTA uses the BERT multilingual tokeniser to generate tokens that BERT's embedding layer will further process. CMTA uses MBERT\footnote{\url{https://github.com/google-research/bert/blob/master/multilingual.md}} to extract contextual features, namely word and sentence embedding vectors, from text data \footnote{Embeddings are helpful for keyword/search expansion, semantic search and information retrieval. They help accurately retrieve results matching a keyword query intent and contextual meaning, even in the absence of keyword or phrase overlap.}. In the subsequent CMTA phases that use NLP models, these vectors are used as feature inputs with several advantages. (M)BERT embeddings are word representations that are dynamically informed by the words around them, meaning that the same word's embeddings will change in (M)BERT depending on its related words within two different sentences.

For the non-expert reader, the tokenization process is based on a WordPiece model. It greedily creates a fixed-size vocabulary of individual characters, subwords, and words that best fit a language data (e.g. English) \footnote{This vocabulary contains whole words, subwords occurring at the front of a word or in isolation (e.g., "em" as in the word "embeddings" is assigned the same vector as the standalone sequence of characters "em" as in "go get em"), subwords not at the front of a word, which are preceded by '\#\#' to denote this case, and individual characters \cite{wu2016google}}. Each token in a tokenized text must be associated with the sentence's index: sentence 0 (a series of 0s) or sentence 1 (a series of 1s). After breaking the text into tokens,   a sentence must be converted from a list of strings to a list of vocabulary indices.
The tokenisation result is used as input to apply BERT that produces two outputs, one pooled output with contextual embeddings and hidden-states of each layer. The complete set of hidden states for this model are stored in a structure containing four elements: the layer number (13 layers) \footnote{It is 13 because the first element is the input embeddings, the rest is the outputs of each of BERT’s 12 layers.}, the batch number (number of sentences submitted to the model), the word / token number in a sentence, the hidden unit/feature number (768 features) \footnote{That is 219,648 unique values to represent our one sentence!}.

In the case of CMTA, the tokenisation is more complex because it is done for sentences written in different languages. Therefore, it relies on the MBERT model that has been trained for this purpose. 


\vspace{-0,35cm}
\paragraph{Feature Extraction Phase}


is intended to exploit the information of hidden-layers produced due to applying BERT to the tokenisation phase result. The objective is to get individual vectors for each token and convert them into a single vector representation of the whole sentence. 
For each token of our input, we have 13 separate vectors, each of length 768. Thus, to get the individual vectors, it is necessary to combine some of the layer vectors. The challenge is to determine which layer or combination of layers provides the best representation.

\vspace{-0,3cm}
\paragraph{Linear convolution}

The hidden states from the 12th layer are processed in this phase, applying linear convolution and pooling to get correlation among tokens.
We apply a three-layer 1D convolution over the hidden states with consecutive pooling layers. The final convolutional layer's output is passed through a global average pooling layer to get a final sentence representation. This representation holds the relation between contextual embeddings of individual tokens in the sentence. 


\vspace{-0,3cm}
\paragraph{Classification}

A linear layer is connected to the model in the end for the CMTA classification task.
This classification layer outputs a Softmax value of vector, depending on the output, the index of the highest value in the vector represents the label for the given sequence: 'false', 'partly false' and 'misleading'.


\subsection{Experiment}\label{sec:experiment}
To validate CMTA, we designed experiments on Google Colab with 64 GB  RAM and 12 GB GPU. For implementing the method, we calibrated pre-trained models provided by hugging face\footnote{\url{https://huggingface.co/}}.
 For our experimental setting, we extracted annotated misinformation data from multiple publicly available open databases. We also collected many multilingual tweets consisting of over 2 million tweets belonging to eight different languages. 
 
\paragraph{Misinformation datasets}
We collected data from an online fact-checker website called Poynter \cite{poynter}. Poynter has a specific COVID-19 related misinformation detection program named 'CoronaVirusFacts/DatosCoronaVirus Alliance Database\footnote{\url{https://www.poynter.org/covid-19-poynter-resources/}}'. This database contains thousands of labelled social media information such as news, posts, claims, articles about COVID-19, which were manually verified and annotated by human volunteers  (fact-checkers) from all around the globe. The database gathers all the misinformation related to COVID-19 cure, detection, the effect on animals, foods, travel, government policies, crime, lockdown. The misinformation dataset is available in 2 languages- `English' and `Spanish'. 

We crawled through the content of  two websites using Beautifulsoup\footnote{\url{https://pypi.org/project/beautifulsoup4/}}, a Python library for scraping information from web pages. We scraped 8471 English language false news/information belonging to nine  classes, namely, 
`False', `Partially false', `Misleading', `No evidence', `Four Pinocchios', `Incorrect', `Three Pinocchios', `Two Pinocchios' and `Mostly False'. We gathered the article's title, its content, and the fact checker's misinformation-type label for each article. 

For  Spanish\footnote{\url{https://chequeado.com/latamcoronavirus/}}, we collected 531 misinformation articles. The collected data contains the misinformation published on social media platforms such as Facebook, Twitter, Whatsapp, YouTube. Posts were mostly related to political-biased news, scientifically dubious information and conspiracy theories, misleading news and rumours about COVID-19. We also used a human-annotated fact-checked tweet dataset \cite{alam2020fighting} available at a public repository\footnote{\url{https://github.com/firojalam/COVID-19-tweets-for-check-worthiness}}. The dataset contained true and false labelled tweets in English and Arabic language. We used only false labelled tweets consisting of 500 English. We compiled  a total of 9,502 micro-articles distributed across 9 misinformation classes shown in Table \ref{tab:misin-table}.

\begin{table}
\small
\centering
\caption{Collected Misinformation Data set.}
\label{tab:misin-table}
\resizebox{0.6\textwidth}{!}{
\begin{tabular}{|l|l|l|}
\hline
\textbf{Classes} & \textbf{Number of tweets}\\
\hline
False \cite{poynter} (English) & 2,869\\
Partially False (English) & 2,765\\
Misleading (English) & 2,837\\
False (Spanish) & 191\\
Partially False (Spanish) & 161\\
Misleading (Spanish) & 179\\
False \cite{alam2020fighting} (English) & 500 \\
\hline
Total & 9,502 \\
\hline
\end{tabular}
}
\end{table}

\paragraph{Dataset Pre-processing}
 The datasets contained noise such as emojis, symbols, numeric values, hyperlinks to websites, and username mentions that were needed to be removed. Since our dataset was multilingual, we had to be very careful while preprocessing as we did not want to lose any valuable information. 
To preprocesses the training and inference datasets,
we used simple regular expressions to remove URLs, special characters or symbols, blank rows, re-tweets, user mentions. We did not remove the hashtags from the data as hashtags might contain helpful information. For example, in the sentence- 'Wear mask to protect yourself from \#COVID-19 \#corona', only the symbol '\#'  was removed.
We removed stop words using NLTK\footnote{NLTK \url{https://www.nltk.org/} is a Python library for natural language processing.}. NLTK supports multiple languages except for few languages, such as Hindi and Thai. For preprocessing the Hindi dataset, we used CLTK (Classical Language Toolkit) \footnote{\url{https://docs.cltk.org/en/latest/index.html}}. For removing Thai stop words from Thai tweets, we used PyThaiNLP \cite{pythainlp}. The emojis were removed using their Unicode. 
\vspace{-0,3cm}
\paragraph{Attribute engineering of the training dataset} 
Table \ref{tab:data}  gives an overview of the training dataset and showcase some misinformation articles.  Column 2 shows the original label assigned by the fact-checker, column 3 gives a misinformation example associated with the label present in column 2, and column 4 provides reasoning given by the fact-checker behind assigning a particular label (column 2) to the misinformation (column 3). For example, if we look at the entry number '3' in the table \ref{tab:data}, the misinformation is about the adverse effect of 5G radiation over the COVID-19 patients. This entry was labelled 'Incorrect' by the fact-checker. After analysing the fact-checker rating and the explanation given, we labelled it as 'False' misinformation. Entry number '5' talks about the COVID-19 test cost. The explanation given by the fact-checker is valid as it is not sure if there is any fee in the USA for the COVID-19 test or not. So because of the lack of evidence and uncertainty, we labelled it as 'Partially false'. Entry number '7' in the table talks about a video showing COVID-19 corpus dumping in the sea. Based on the explanation, the video was coupled with the wrong information to mislead the audience. So it was labelled as 'Misleading' misinformation.

The collected data is unevenly distributed across nine classes: 'No evidence', 'Four Pinocchios\footnote{90\%-95\% changes of it being false}', 'Incorrect', 'Three Pinocchios\footnote{70\%-75\% changes of it being false}', 'Two Pinocchios\footnote{50\%-55\% changes of it being false}' and 'Mostly False' (the smallest group).  Most collected articles were labelled either as 'False', 'Partially false' and 'Misleading'. 
We performed an attribute engineering phase for preparing the dataset. 
 We produced a  uniformly distributed dataset reorganised the initial dataset as follows.   The classes 'Four Pinocchios' and 'Incorrect' were merged with the class 'False'. The classes 'Three Pinocchios' and 'Two Pinocchios' were merged into the class 'Partially false'. The classes 'No evidence' and 'Mostly False' were merged with the class 'Misleading'. 
Finally, column 1 (see table \ref{tab:data}) corresponds to the label assigned during the attribute engineering phase.


\begin{table*}[t]
\caption{Misinformation Dataset} \label{tab:data}
\resizebox{\textwidth}{!}{
\begin{tabular}{|cc|cl|ll|l|}
\hline

\textbf{Our Rating}                                                                 & & \textbf{IFCN(Poynter) Rating}      &  & \textbf{Misinformation }                                                                                                                                                                     &  & \textbf{Explanation}                                                                                                                                                                                                                                                                                                                                                                                                                                                                                                                                                                                         \\ \cline{1-1} \cline{3-3} \cline{5-5} \cline{7-7} 
False                                                     & & False            & & \begin{tabular}[c]{@{}l@{}}The border between France \\and Belgium will be closed.\end{tabular}                                                                                 & & \begin{tabular}[c]{@{}l@{}}French and Belgian authorities\\denied it.\end{tabular}                                                                                                                                                                                                                                                                                                                                                                                                                \\ \cline{3-3} \cline{5-5} \cline{7-7} 
                                                                           & & Four pinocchios  & & \begin{tabular}[c]{@{}l@{}}Trump’s effort to blame Obama\\ for sluggish coronavirus testing.\end{tabular}                                                                  & & \begin{tabular}[c]{@{}l@{}}There was no “Obama rule,” just\\draft guidance that never took\\effect and was withdrawn before\\President Trump took office.\end{tabular}                                                                                                                                                                                                                                                                                                                                                                                                                               \\ \cline{3-3} \cline{5-5} \cline{7-7} 
                                                                           & & Inaccurate        & & \begin{tabular}[c]{@{}l@{}}Elisa Granato, the first volunteer\\in the first Europe human tria\\of a COVID-19 vaccine, has died.\end{tabular} &  & \begin{tabular}[c]{@{}l@{}}Elisa Granato, the first volunteer\\in the first Europe human trial\\of a COVID-19 vaccine, has died.\end{tabular} \\ \hline
Partially False & & Partially False  & & \begin{tabular}[c]{@{}l@{}}Media shows a Florida beach\\ full of people while it’s empty.\end{tabular}                                                                     & & \begin{tabular}[c]{@{}l@{}}The different videos were not shot\\at the same time. The beaches\\are empty when they are closed.\end{tabular}                                                                                                                                                                                                                                                                                                                                                                                                                                                            \\ \cline{3-3} \cline{5-5} \cline{7-7} 
                                                                           & & Two Pinocchios     & & \begin{tabular}[c]{@{}l@{}}The bill for a coronavirus\\ test in the US is \$3.000\end{tabular}                                                                             & & \begin{tabular}[c]{@{}l@{}}The CDC is not making people\\pay the test by now.\end{tabular}                                                                                                                                                                                                                                                                                                                                                                                                                                                                                                             \\ \cline{3-3} \cline{5-5} \cline{7-7} 
                                                                           & & Partly False & & \begin{tabular}[c]{@{}l@{}}Salty and sour foods cause\\the “body of the COVID-19 virus"\\to explode and dissolve.\end{tabular}                                                         & & \begin{tabular}[c]{@{}l@{}}“Consuming fruit juices or gargling\\with warm water and salt does not\\protect or kill COVID-19,” the\\World Health Organization\\Philippines told VERA Files.\end{tabular}                                                                                                                                                                                                                       \\ \hline
Misleading                                               & & Misleading       & & \begin{tabular}[c]{@{}l@{}}A clip from Mexico depicts\\ the dumping of coronavirus\\ patients corpses into the sea.\end{tabular}                                           & & \begin{tabular}[c]{@{}l@{}}Misbar's investigation of the video\\revealed that it does not depict the\\dumping of coronavirus patients\\corpses in Mexico, but rather paratroopers\\landing from a Russian MI 26 helicopter.\end{tabular}                                                                                                                                                                                                                                                                                                                                                            \\ \cline{3-3} \cline{5-5} \cline{7-7} 
                                                                           & & No Evidence      & & \begin{tabular}[c]{@{}l@{}}Media uses photos of puppets on\\patient stretchers to scare the\\public.\end{tabular}                                                          & & \begin{tabular}[c]{@{}l@{}}There is no evidence that any media\\ outlet used this photo for their reporting\\ about COVID-19. Its origin is unclear,\\ maybe it was shot in Mexico and shows\\ a medical training session.\end{tabular}                                                                                                                                                                                                                                                                                                                                                                 \\ \cline{3-3} \cline{5-5} \cline{7-7} 
                                                                           & & Mostly False     & & \begin{tabular}[c]{@{}l@{}}Coronavirus does not affect\\ people with ‘O+’ blood type.\end{tabular}                                                                         & & \begin{tabular}[c]{@{}l@{}}The post claiming coronavirus does\\ not affect people with ‘O+’ blood\\ type is misleading.\end{tabular}                                                                                                                                                                                                                                                                                                                                                                                                                                                                    \\ \hline
\end{tabular}
}
\end{table*}

\vspace{-0,35cm}
\paragraph{{ Inference Dataset}}
For building an inference dataset to be used to test the CMTA trained model for analysing the misinformation spread across all over the social media platforms in multiple languages,
We collected around 2,137,106 multilingual tweets. The tweets were expressed  in eight major languages, namely- 'English', 'Spanish', 'Indonesian', 'French', 'Japanese', 'Thai', 'Hindi' and 'German'. Therefore, we used a dataset of tweets IDs associated with the novel coronavirus COVID-19 \cite{chen2020tracking}. Starting on January 28, 2020, the current dataset contains 212,978,935 tweets divided into groups based on their publishing month. The dataset was collected using multilingual COVID-19 related keywords and contained tweets in more than 30 languages. We used tweepy\footnote{Python module is available at \url{http://www.tweepy.org}{http://www.tweepy.org}} which is a Python module for accessing Twitter API. We decided to retrieve the tweets using the tweet IDs published in the past five months (February, March, April, May and June) for our analysis. Table \ref{month-table} shows the total number of collected tweets. The distribution of tweets across eight languages corresponds to most English items, almost 1 and 1/8 of the whole data set, then Spanish (1/4 of the total number of tweets) and the rest for French, Japanese, Indonesian, Thai, and Hindi.
\vspace{-0,3cm}
\begin{table}
\centering
\small
\caption{\label{month-table} Language-wise Dataset Distribution}
\resizebox{0.5\textwidth}{!}{
\begin{tabular}{|l|c|c|}
\hline \textbf{Language}& \textbf{ISO} & \textbf{Number of tweets} \\ \hline
English	& en & 1,472,448\\
Spanish	& es & 353,294\\
Indonesian & in & 80,764\\
French & fr & 71,722\\
Japanese & ja & 71,418\\
Thai & th & 36,824\\
Hindi & hi & 27,320\\
German & de & 23,316 \\
\hline
Sum & & 2137106\\
\hline
\end{tabular}
}
\end{table}


\subsection{Model Setup and Training}
\label{mis_model}

\paragraph{Training Setting}

For training our model, we divided the data into training, validation and testing datasets in the ratio of 80\%/10\%10\% respectively. The final count for the train, validation and test dataset was 7,602, 950, 950.
We fine-tuned the Sequence Classifier from HuggingFace based on the parameters specified in \cite{devlin2018bert}. Thus, we set a batch size of 32, learning rate 1e-4, with Adam Weight Decay as the optimizer. We run the model for training for 10 epochs. Then, we save the model weights of the transformer, helpful for further training.

\vspace{-0,3cm}
\paragraph{Hyperparameters' Setting}
Table \ref{parameter} lists every hyperparameter for training and testing our model. All the calculations and selection of hyperparameters were made based on tests and the model's best output. After performing several iterations on distinct sets of hyper-parameters based on the model's performance analysis, we adopted the one showing promising results on our dataset.
\vspace{-0,3cm}
\begin{table}
\centering
\small
\caption{\label{parameter}Hyper-parameters for training}
\resizebox{0.5\textwidth}{!}{
\begin{tabular}{|l|l|l|}
\hline \textbf{Parameters}& \textbf{Value} \\  \hline
Pool Size of Average Pooling            &  8                                  \\
Pool Size of Max Pooling                 & 8                                  \\
Dropout Probability                       & 0.36                               \\
Number of Dense layers                  & 4                                  \\
Text Length                             & 128                                \\
Batch Size                               & 32                                 \\
Epochs                                  & 10                                 \\
Optimizer                               & Adam                               \\
Learning Rate                           & $1\times10^{-4}$   \\
\hline
\end{tabular}
}
\end{table}

\subsection{{Experiment and Results}}
\begin{figure}
\centering
\includegraphics[width=0.6\textwidth]{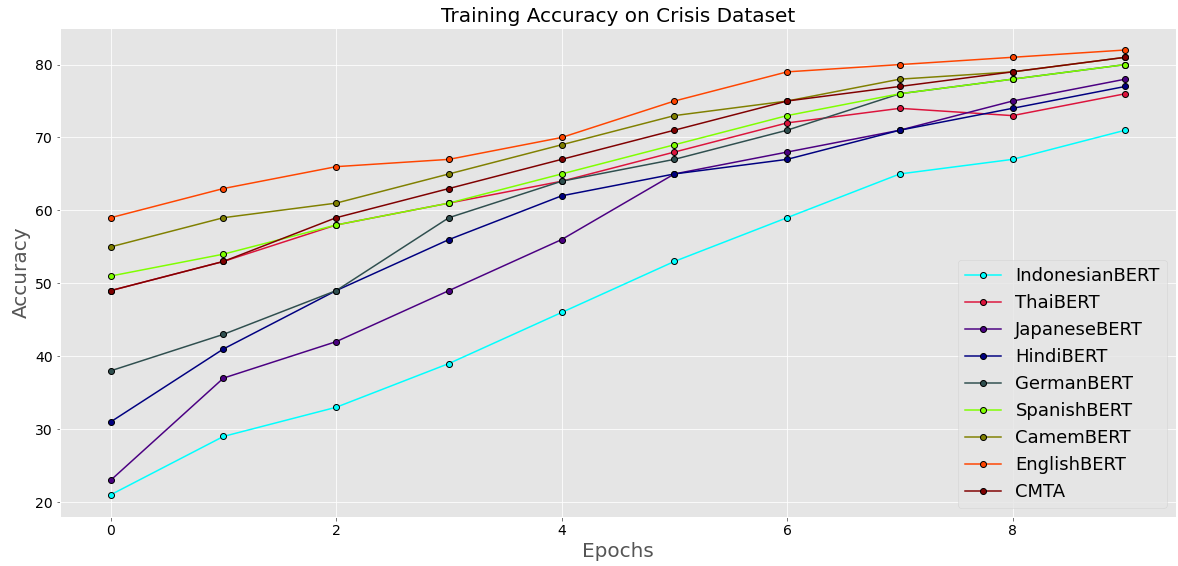} \label{fig:1a}
\includegraphics[width=0.6\textwidth]{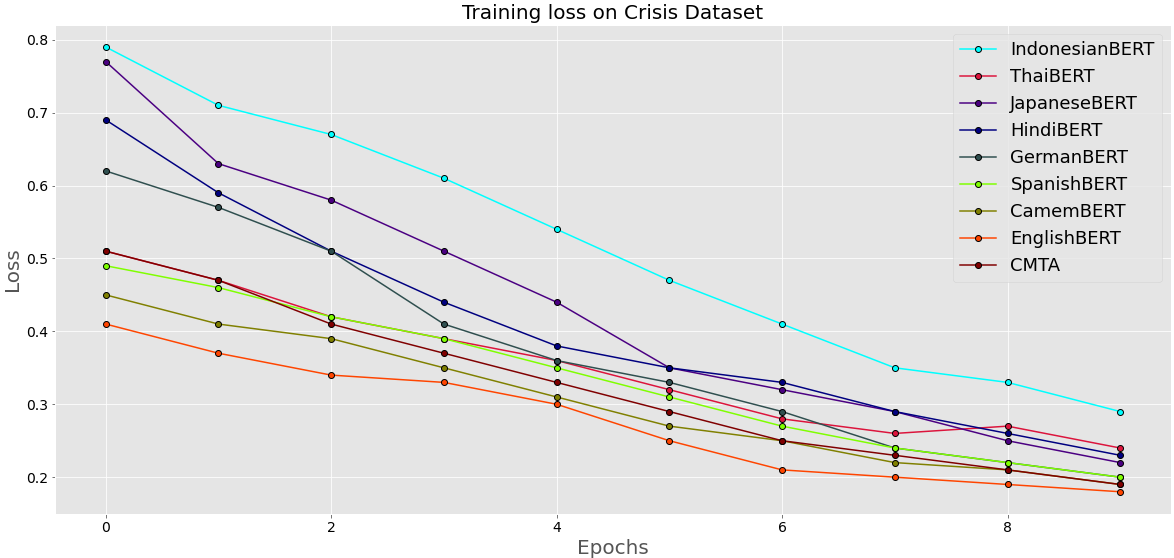}\label{fig:1b}
  \caption{Training Accuracy(Upper) and Training loss(Lower)} \label{fig:accuracy}
 \end{figure}

We experimented with the multilingual data with their respective linguistic-based BERT models. We set the model with the same training parameters as the CMTA model and preprocessed the data as stated previously. Each monolingual model was fine-tuned for 10 epochs with a batch size of 32, and it was applied to the classification dataset of their respective language. 
Our model achieved an accuracy(\%) of \textbf{82.17} (see figure \ref{fig:accuracy})and $F_1$ score (\%) of \textbf{82.54} on the test dataset. The precision and recall reported by the model were \textbf{82.07} and \textbf{82.30} respectively.

 Table \ref{result} shows the model's prediction over few examples from the test dataset along with their actual label. As we have shown in the table,  in the entry numbers '1',  '2', '3' and '4' our model could predict the correct label. However, in the case of entry number '5', the label predicted by our model was 'False', whereas the actual label is 'Misleading'. If we would look at the misinformation at the entry number '5', which is a Spanish text- 'El medicamento contra piojos sirve como tratamiento contra Covid-19.' and the corresponding English translation would be- ''. This misinformation claims about a COVID-19 medicine, and since this could be 'false' and 'misleading' misinformation at the same time, our model predicted it as 'false' misinformation rather than 'misleading'. 

\begin{table}
\centering
\small
\caption{\label{result}Misinformation data examples along with model's prediction and actual label}
\resizebox{\textwidth}{!}{
\begin{tabular}{|l|l|l|c|}
\hline \textbf{Test Data}& \textbf{Actual Label} & \textbf{Prediction} & \textbf{Accuracy(\cmark/\xmark)} \\  \hline
Dr. Megha Vyas from India died due to & & &  \\
COVID-19 while treating COVID patients.&   False   & False   & \cmark   \\
El plátano bloquea “la entrada celular& & &  \\ del COVID-19”                                 & False                                     & False                                   & \cmark                   \\
Asymptomatic people are very rarely & & &  \\ contagious, said the WHO.                     & Partially False                             & Partially False                          & \cmark                   \\
Patanjali Coronil drops can help cure & & &  \\ coronavirus.                                                                            & Misleading                               & Misleading                           & \cmark                   \\
El medicamento contra piojos sirve como & & &  \\ tratamiento contra Covid-19.            & Misleading                          & False                                 & \xmark                   \\
\hline
\end{tabular}
}
\end{table}

\section{Results Assessment }\label{sec:assessment}
We used two strategies for assessing CMTA results according to the research questions we wanted to prove. The first research question was to determine (Q$_1$) whether it was possible to develop a method that could provide a general multi-lingual classification pipeline? The second question was (Q$_2$) whether it is possible to build conclusions about how misinformation on COVID-19 spreads in different language speaking communities by analysing micro-texts published in social media. For Q$_1$, we conducted a comparative study of CMTA with different independent mono-lingual misinformation classifiers. Thereby, CMTA's classification performance for a given set of micro-texts written in a given language was compared against a classification model targeting only that language. For 
Q$_2$ we analysed and plotted CMTA's results, and we proposed intuitive arguments according to our observations. 

\subsection{CMTA vs Monolingual Classification}
We conducted a comparative performance study of various monolingual BERT models concerning our proposed multilingual CMTA model for comparing their performance for the misinformation detection task. We investigated eight monolingual BERT model\footnote{Pretrained model available at \url{https://huggingface.co/models}}, namely, 'English', 'Spanish', 'French', 'Germann', 'Japanese', 'Hindi' 'Thai\footnote{ThaiBERT is available at \url{https://github.com/ThAIKeras/bert}}' and 'Indonesian'.
%
We used the same 9,502 tweets distributed across three misinformation classes for training the monolingual models. Our dataset consisted of English and Spanish tweets; therefore, we translated the tweets into eight languages to train each of the eight monolingual models. We used Google Translator API\footnote{Please refer \url{https://cloud.google.com/translate/docs}} for converting the tweets into a particular language. Results are shown in Table \ref{Metrics}.
Based on the experiment results, we strongly suggest that the multilingual CMTA model could generalize smoothly on the dataset because its performance was equivalent to the monolingual models.


\vspace{-0,3cm}
\begin{table}
\centering
\small
\caption{\label{Metrics}Precision, Recall and f-score of CMTA
model}
\resizebox{0.6\textwidth}{!}{
\begin{tabular}{|l|c|c|c|}
\hline
\textbf{\diagbox{Models}{Metrics}} &          \textbf{Precision} & \textbf{Recall} & \textbf{F1-score} \\ \hline
EnglishBERT                                            & 82.03              & 74.18           & 77.90             \\
SpanishBERT                                           & 80.9               & 72.02           & 76.20             \\
CamemBERT                                             & 81.91              & 71.45           & 76.32             \\
GermanBERT                                             & 80.61              & 71.43           & 75.74             \\
JapaneseBERT                                            & 79.56              & 65.36           & 71.76             \\
HindiBERT                                              & 79.56              & 65.68           & 71.95             \\
ThaiBERT                                               & 79.11              & 66.25           & 72.11             \\
IndonesianBERT                                         & 78.96              & 65.66           & 71.69             \\
CMTA                                                    & 81.52              & 74.40           & 77.79             \\ 
\hline
\end{tabular}
}
\end{table}

\subsection{Multilingual Misinformation Analysis}\label{sec:multilingual-analysis}
We provide a detailed analysis of misinformation distribution across multilingual tweets. This analysis responds to the initial question: {\em how is misinformation about COVID-19 spread in communities speaking different languages.} Our survey studied and analyzed the distribution of COVID-19 misinformation across eight significant languages (i.e. 'English', 'Spanish', 'Indonesian', 'French', 'Japanese', 'Thai', 'Hindi' and 'German') for five months (i.e. February, March, April, May and June). 

\begin{figure}
\centering
\includegraphics[width=\linewidth]{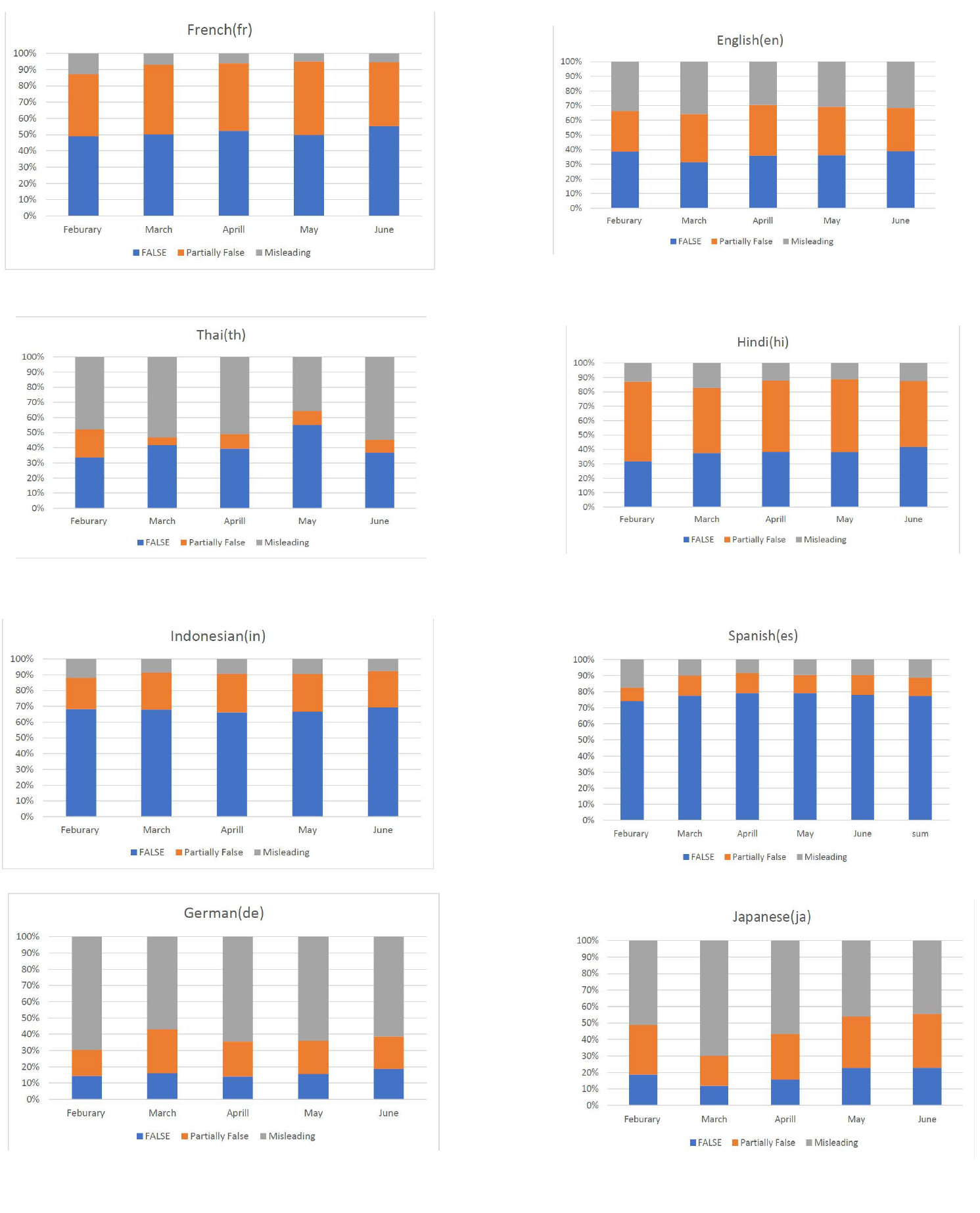}
\caption{Month-wise Disinformation Distribution in Languages.}
\label{fig:lang_distribution_monthly}
\end{figure}

We used our trained, multilingual model, CMTA, to predict and categorize the misinformation type present in tweets.  We conducted our sequential misinformation analysis on a collection of over 2 million multilingual tweets. Figure \ref{fig:lang_distribution_monthly} shows the month-wise distribution of misinformation types for each language. It showcases the overall (all 5 months together) spread of misinformation types across each language. We could see that German tweets have the highest number of 'Misleading' tweets, whereas French have the least. Spanish tweets beat other language's tweets by becoming the largest source of 'False' misinformation. Germany generated the least number of 'False' tweets. Hindi tweets tend to have the highest number of 'Partially false' tweets, whereas Thai have the least. 

\begin{figure}
\centering
\includegraphics[width=0.6\textwidth]{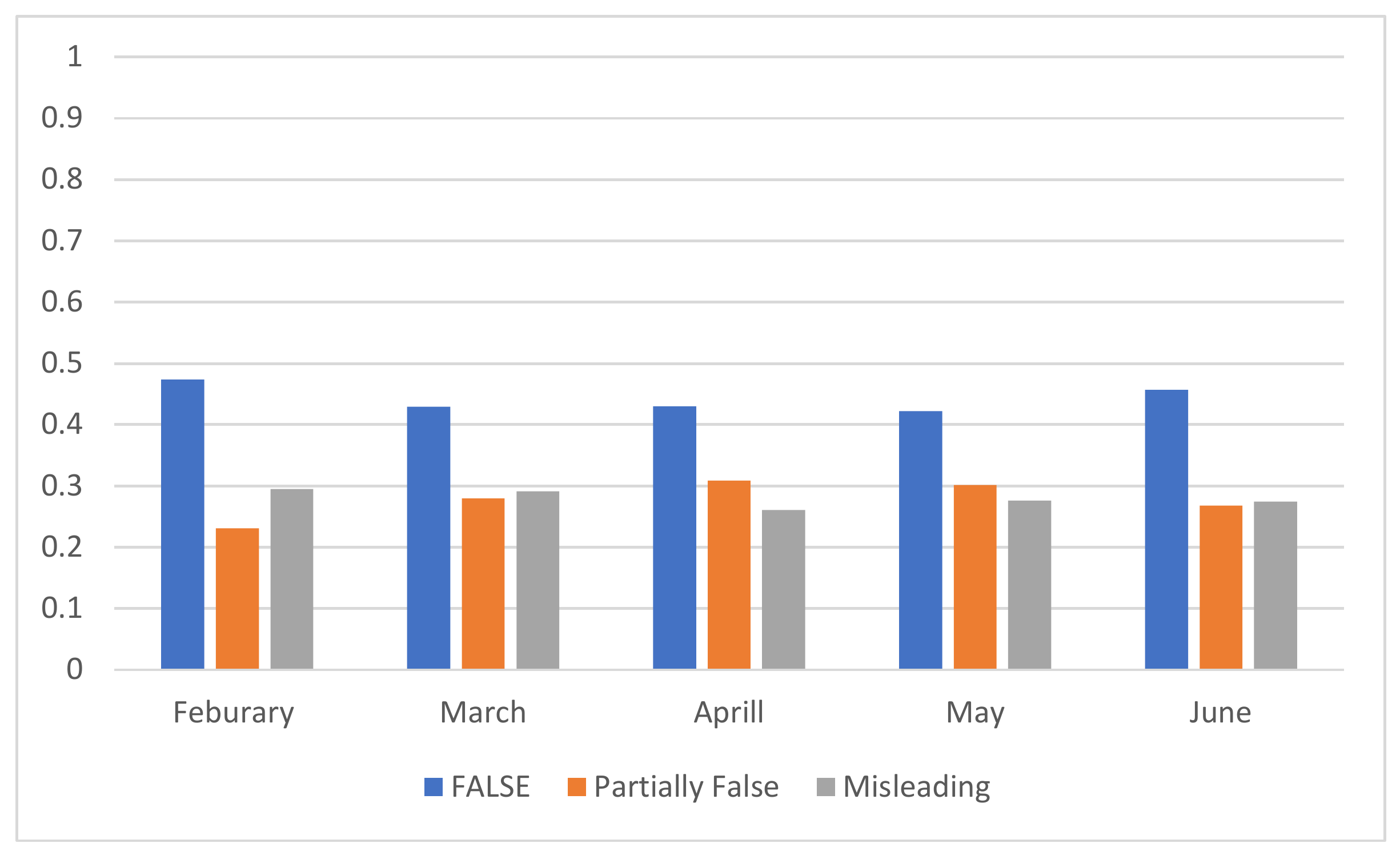}
\caption{Month-wise Disinformation Distribution.}
\label{FIG:month_distribution}
\end{figure}

We could observe that for February, March and June months, our model predicted a large number of tweets as 'False', followed by 'Misleading', which is the second largest and the number of 'Partially false' was the least (see Figure \ref{FIG:month_distribution}). Our model discovered that the number of 'Partially false' tweets are more than 'Misleading' tweets and 'False' tweets were again in the majority for the tweets generated during April and May.

\begin{table}[t]
\centering
\small
\caption{\label{full_res}Language-wise predicted misinformation labels of tweets in February, March and April.}
\resizebox{\textwidth}{!}{
\begin{tabular}{|ll|llll|llll|lll|}
\hline
\textbf{Lingo} &  & \multicolumn{3}{|c|}{\textbf{February}}                                                                                       &  & \multicolumn{3}{|c|}{\textbf{March}}                                                                                          &  & \multicolumn{3}{|c|}{\textbf{April}}                                                                                          \\ \cline{3-5} \cline{7-9} \cline{11-13} 
\multicolumn{1}{|c}{}                                &  & \multicolumn{3}{|c|}{\textbf{Misinformation}}                                                                                 &  & \multicolumn{3}{|c|}{\textbf{Misinformation}}                                                                                 &  & \multicolumn{3}{|c|}{\textbf{Misinformation}}                                                                                 \\ \cline{3-5} \cline{7-9} \cline{11-13} 
\multicolumn{1}{|c}{}                                &  & \multicolumn{1}{|c|}{\textbf{False}} & \multicolumn{1}{c}{\textbf{Partially False}} & \multicolumn{1}{|c|}{\textbf{Misleading}} &  & \multicolumn{1}{|c|}{\textbf{False}} & \multicolumn{1}{c}{\textbf{Partially False}} & \multicolumn{1}{|c|}{\textbf{Misleading}} &  & \multicolumn{1}{|c|}{\textbf{False}} & \multicolumn{1}{c}{\textbf{Partially False}} & \multicolumn{1}{|c|}{\textbf{Misleading}} \\ \cline{1-1} \cline{3-5} \cline{7-9} \cline{11-13} 
\textbf{Spanish}                                    &  & 58346                              & 6653                                         & 13740                                   &  & 67956                              & 10913                                        & 8826                                    &  & 34125                              & 5437                                         & 3604                                    \\
\textbf{German}                                     &  & 517                                & 581                                          & 2505                                    &  & 862                                & 1438                                         & 3043                                    &  & 584                                & 892                                          & 2664                                    \\
\textbf{Japanese}                                   &  & 1920                               & 3079                                         & 5245                                    &  & 448                                & 692                                          & 2650                                    &  & 1635                               & 2850                                         & 5840                                    \\
\textbf{Indonesian}                                 &  & 11157                              & 3226                                         & 1951                                    &  & 12573                              & 4336                                         & 1582                                    &  & 9073                               & 3367                                         & 1273                                    \\
\textbf{English}                                    &  & 88369                              & 62747                                        & 76640                                   &  & 92428                              & 96571                                        & 105143                                  &  & 77368                              & 74947                                        & 63473                                   \\
\textbf{French}                                     &  & 4464                               & 3472                                         & 1155                                    &  & 12024                              & 10270                                        & 1670                                    &  & 6650                               & 5300                                         & 763                                     \\
\textbf{Hindi}                                      &  & 500                                & 870                                          & 202                                     &  & 756                                & 909                                          & 348                                     &  & 2211                               & 2868                                         & 705                                     \\
\textbf{Thai}                                       &  & 1950                               & 1074                                         & 2780                                    &  & 6036                               & 736                                          & 7678                                    &  & 2263                               & 554                                          & 2917                                    \\ \hline
\end{tabular}
}
\end{table}

\begin{table}[t]
\centering
\small
\caption{\label{full_res2} Language-wise predicted misinformation labels of tweets in May and June.}

\resizebox{0.9\textwidth}{!}{
\begin{tabular}{|ll|llll|lll|}
\hline
Lingo &  & \multicolumn{3}{|c|}{\textbf{May}}                                                                                            &  & \multicolumn{3}{|c|}{\textbf{June}}                                                                                           \\ \cline{3-5} \cline{7-9} 
\multicolumn{1}{|c}{}                                &  & \multicolumn{3}{|c|}{\textbf{Misinformation}}                                                                                 &  & \multicolumn{3}{|c|}{\textbf{Misinformation}}                                                                                 \\ \cline{3-5} \cline{7-9} 
\multicolumn{1}{|c}{}                                &  & \multicolumn{1}{|c}{\textbf{False}} & \multicolumn{1}{|c|}{\textbf{Partially False}} & \multicolumn{1}{|c|}{\textbf{Misleading}} &  & \multicolumn{1}{|c|}{\textbf{False}} & \multicolumn{1}{|c|}{\textbf{Partially False}} & \multicolumn{1}{|c|}{\textbf{Misleading}} \\ \cline{1-1} \cline{3-5} \cline{7-9} 
\textbf{Spanish}                                    &  & 57821                              & 8214                                         & 7107                                    &  & 54965                              & 8828                                         & 6759                                    \\
\textbf{German}                                     &  & 1076                               & 1426                                         & 4430                                    &  & 616                                & 657                                          & 2028                                    \\
\textbf{Japanese}                                   &  & 8984                               & 12324                                        & 18125                                   &  & 1741                               & 2496                                         & 3389                                    \\
\textbf{Indonesian}                                 &  & 12695                              & 4574                                         & 1805                                    &  & 9114                               & 3038                                         & 1000                                    \\
\textbf{English}                                    &  & 140494                             & 128326                                       & 119391                                  &  & 135172                             & 101896                                       & 109483                                  \\
\textbf{French}                                     &  & 8475                               & 7667                                         & 842                                     &  & 4952                               & 3535                                         & 483                                     \\
\textbf{Hindi}                                      &  & 4560                               & 6057                                         & 1343                                    &  & 2501                               & 2739                                         & 751                                     \\
\textbf{Thai}                                       &  & 2825                               & 470                                          & 1830                                    &  & 2103                               & 486                                          & 3122                                    \\ \hline
 \end{tabular}
 }
 \end{table}

Table \ref{full_res} and \ref{full_res2} present a detailed count of misinformation classes across all the languages. 
The following  specific observations were made concerning the languages:  
%
The misinformation distribution for English data indicates that there is a majority of \textbf{False} tweets during the five months, whereas the distribution of \textbf{Misleading} labelled data is slightly less than as compared to \textbf{False} labelled data. \textbf{Partially False} labelled tweets are moderately distributed, as in the month April, we can see that there is a more significant number concerning other months.
    %
    According to the language wise-distribution shown in Figure \ref{fig:lang_distribution_monthly}, Spanish tweets have a greater frequency of \textbf{False} labelled tweets, whereas the \textbf{Misleading} tweets and \textbf{Partially False} tweets shows the almost identical number of tweet across the five months. 
    %
    There was a surge of \textbf{Misleading} labelled tweets during February, and the count remained the same throughout the five months. There was also an increase in \textbf{Partially False} tweets in March, but it decreased in successive months, leading to minor \textbf{False} labelled tweets.
%
   According to the language wise-distribution shown in Figure \ref{fig:lang_distribution_monthly},  on average throughout the five months, approx 20\% of Japanese tweets are labelled \textbf{False}. Similarly, approx 30\% of the Japanese tweets are labelled \textbf{Partially False}, leading to the majority of 50\% data are labelled as \textbf{Misleading}. We can also see a considerable increase in \textbf{Misleading} tweets in March, tweeted in the Japanese language.
    %
    According to the distribution of Indonesian tweets, approximately 10\% of tweets are labelled as \textbf{Misleading}, and on the contrary, there is a large distribution of \textbf{False} labelled tweets. Approximately 34\% of the Indonesian dialect data is labelled as \textbf{Partially False} throughout the five months.
    %
    %
    %
    Figure\ref{fig:lang_distribution_monthly} shows the misinformation distribution across all five months in the French tweets. The largest majority of the tweets were classified as \textbf{False} misinformation. Among \textbf{Partially false} and \textbf{Misleading}, the least number of tweets were labelled as \textbf{Misleading}. 
    %
    The frequency of Hindi tweets is low in the dataset used in our experiment. However, our model can predict or label Hindi tweets. Tweets in Hindi have low numbers of \textbf{Misleading} tweets, whereas the \textbf{Partially False} tweets class has a great frequency. \textbf{False} labelled tweets are slightly low compared to \textbf{Partially False} tweets in this dialect.
    %
    The distribution of Thai tweets shows that our model prediction is majorly oriented towards the \textbf{Misleading} tweets. The distribution of \textbf{Misleading} labelled tweets is the greatest among the labelled classes, in contrast to \textbf{Partially False} tweets. \textbf{False} labelled tweets are comparatively moderate in this language.
    

\section{Conclusion and Future Work} \label{sec:conclusion}
This paper introduced CMTA, a multilingual model for analyzing text applied to classify COVID-19 related multilingual tweets into misinformation categories. We demonstrated that our multilingual CMTA framework performed significantly well compared to the monolingual misinformation detection models used independently.
Experimental validation of CMTA detected misinformation distribution across eight significant languages. The paper presented a quantified magnitude of misinformation distributed across different languages in the last five months.  
Our future work aims to collect more annotated training data and perform analysis of a larger multilingual dataset to gain a deeper understanding of misinformation spread. We are currently improving our model's robustness and contextual understanding for better performance in the classification task. We hope that researchers could gain deeper insights about misinformation spread across major languages and use the information to build more reliable social media platforms through our work. 
%
%
%
%

\bibliographystyle{splncs04}
\bibliography{cas-refs}

\end{document}